\documentclass[letterpaper]{article} 
\usepackage{aaai2026}  

\usepackage{times}  
\usepackage{helvet}  
\usepackage{courier}  
\usepackage[hyphens]{url}  
\usepackage{graphicx} 
\usepackage{amsmath}    
\usepackage{amssymb}   
\usepackage{booktabs}  
\usepackage{xcolor} 
\usepackage{natbib}  
\usepackage{caption} 
\usepackage{pifont}
\usepackage{multirow}
\usepackage{algorithm}
\usepackage{algorithmic}
\usepackage{enumitem}
\usepackage{newfloat}
\usepackage{listings}
\DeclareCaptionStyle{ruled}{labelfont=normalfont,labelsep=colon,strut=off} 
\floatstyle{ruled}
\newfloat{listing}{tb}{lst}{}
\floatname{listing}{Listing}

\title{HyperCOD: The First Challenging Benchmark and Baseline for \\ Hyperspectral Camouflaged Object Detection}

\author{
    Shuyan Bai,
    Tingfa Xu*,
    Peifu Liu,
    Yuhao Qiu,
    Huiyan Bai,
    Huan Chen,
    Yanyan Peng,
    Jianan Li*
}
\affiliations{
School of Optics and Photonics, Beijing Institute of Technology, \\
    5 South Zhongguancun Street, Haidian District, Beijing 100081, China \\
}

\usepackage{bibentry}

\begin{document}

\maketitle

\begin{abstract}
RGB-based camouflaged object detection struggles in real-world scenarios where color and texture cues are ambiguous. While hyperspectral image offers a powerful alternative by capturing fine-grained spectral signatures, progress in hyperspectral camouflaged object detection (HCOD) has been critically hampered by the absence of a dedicated, large-scale benchmark. To spur innovation, we introduce HyperCOD, the first challenging benchmark for HCOD. Comprising 350 high-resolution hyperspectral images, It features complex real-world scenarios with minimal objects, intricate shapes, severe occlusions, and dynamic lighting to challenge current models.
The advent of foundation models like the Segment Anything Model (SAM) presents a compelling opportunity. To adapt the Segment Anything Model (SAM) for HCOD, we propose HyperSpectral Camouflage-aware SAM (HSC-SAM). HSC-SAM ingeniously reformulates the hyperspectral image by decoupling it into a spatial map fed to SAM's image encoder and a spectral saliency map that serves as an adaptive prompt. This translation effectively bridges the modality gap. Extensive experiments show that HSC-SAM sets a new state-of-the-art on HyperCOD and generalizes robustly to other public HSI datasets. The HyperCOD dataset and our HSC-SAM baseline provide a robust foundation to foster future research in this emerging area.

\end{abstract}
%
\begin{links}
    \link{Code}{https://github.com/Baishuyanyan/HyperCOD}
    \link{Datasets}{https://github.com/Baishuyanyan/HyperCOD}
\end{links}

\section{Introduction}

\begin{figure}[t]
  \includegraphics[width=\linewidth]{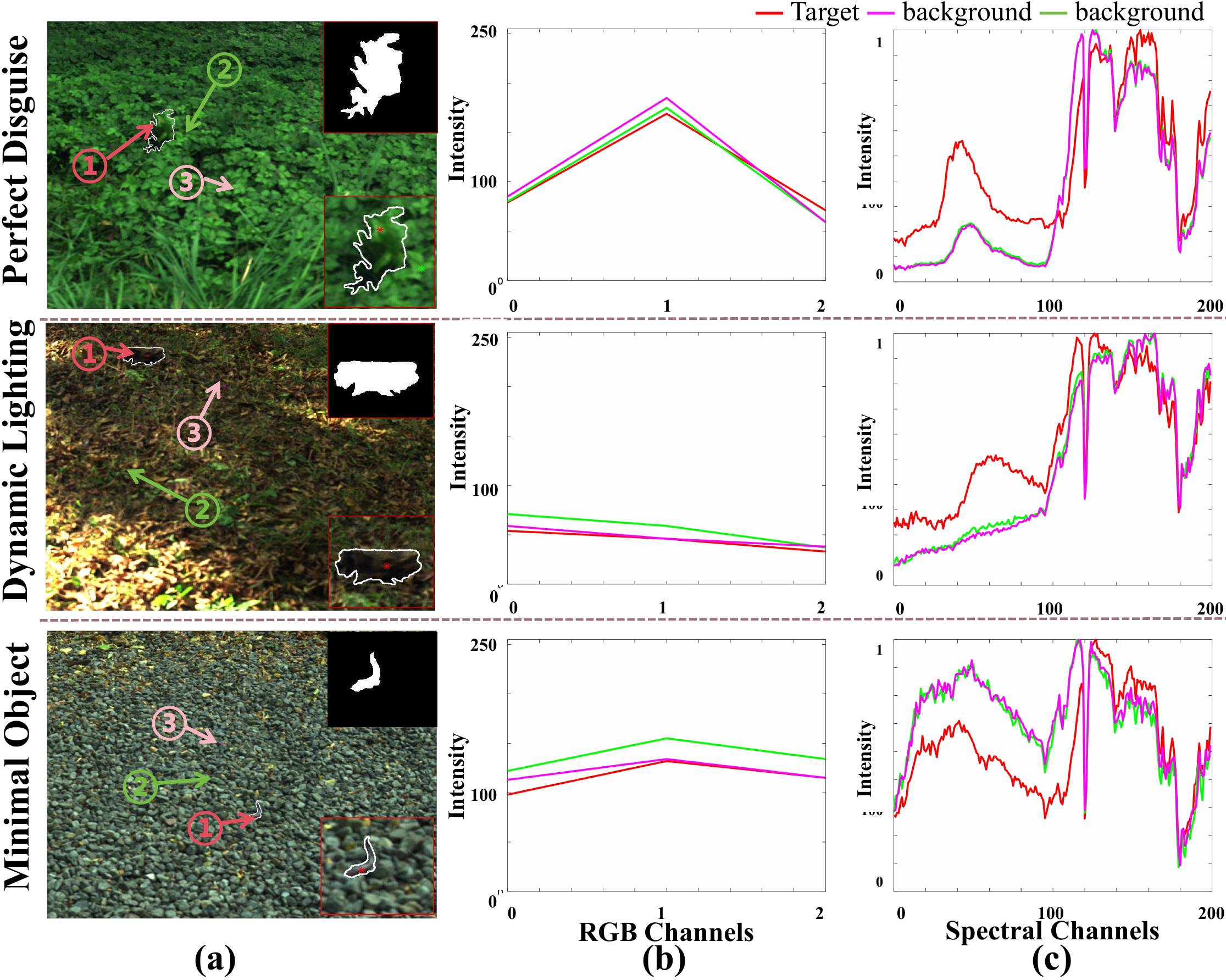}
  \caption{While RGB images show near-indistinguishable camouflage, hyperspectral signatures reveal significant differences, highlighting spectral advantages for COD.}
  \label{fig:motivation}
\end{figure}

Camouflaged object detection (COD) identifies objects blending with their surroundings~\shortcite{paper2,paper3}, with applications in medical diagnosis~\shortcite{paper38}, ecological monitoring~\shortcite{paper39} and search-and-rescue~\shortcite{paper40}. Existing RGB-based methods rely on spatial cues like texture and edges, but fail when camouflage is perfected or illumination flattens spatial distinctions (Fig.~\ref{fig:motivation}(a)(b)), exposing limitations of color-based vision.

Hyperspectral image (HSI) captures hundreds of spectral bands, revealing material differences invisible to RGB cameras~\shortcite{10852603} (Fig.~\ref{fig:motivation}(c)). The unique spectral properties of HSI make it exceptionally well-suited for detecting camouflaged objects, especially those that are small, partially occluded, or set against cluttered backdrops. We formalize this capability as \textbf{Hyperspectral Camouflaged Object Detection (HCOD)}, a novel task currently constrained by the lack of large-scale, diverse benchmark datasets.

To catalyze research and provide a standardized evaluation platform for HCOD, we present \textbf{HyperCOD}, the first comprehensive benchmark for hyperspectral camouflaged object detection. HyperCOD features 350 high-quality HSI cubes, each with high spatial resolution ($1240 \times 1680$) and dense spectral information (200 bands from 400-1000 nm). Captured across 11 diverse natural environments (grasslands, snowfields, etc.), the dataset incorporates challenging scenarios including small targets, complex occlusions, cluttered backgrounds, and dynamic lighting.  It is designed not only to benchmark existing methods but also to drive the development of novel algorithms that can fully exploit spectral information for robust real-world camouflage detection.

While hyperspectral image provides rich information, effectively harnessing it remains a challenge. Traditional hyperspectral analysis methods often rely on pixel-wise, hand-crafted spectral metrics~\shortcite{paper45}, which lack spatial context and exhibit poor robustness in complex scenes~\shortcite{paper28}. On the other hand, the remarkable success of large vision models like the Segment Anything Model (SAM) in RGB domains is inspiring. However, their direct application to hyperspectral images is infeasible due to the prohibitive computational cost and the modality gap between RGB and high-dimensional hyperspectral images.

To bridge this gap, we propose \textbf{HSC-SAM}, a novel framework designed to adapt the powerful segmentation capabilities of SAM for the HCOD task. To resolve the modality mismatch, our Spectral-Spatial Decomposition Module (SSDM) decouples HSI into two complementary representations: a CIEXYZ color-space mapped image that preserves spatial structures and a spectral saliency map that highlights the camouflaged target. This design allows SAM's RGB-pretrained encoder to process familiar spatial structures while using the spectral map as a high-quality, spectrum-driven prompt to guide segmentation.

To improve computational efficiency, our Spectral-Guided Token Dropout (SGTD) mechanism leverages the generated saliency map to selectively discard image tokens from non-salient background regions prior to the expensive attention computations. This principled filtering reduces the processing load and focuses the model's capacity on promising foreground areas. Besides, the Fusion Detail Enhancer (FDE) re-injects high-frequency spatial features to refine object contours and ensure precise boundary localization.

Experiments on HyperCOD validate HSC-SAM's state-of-the-art performance (MAE: 0.0017) and its effectiveness on hyperspectral salient object detection. Our HyperCOD dataset and the HSC-SAM framework provide a solid foundation for future advancements in hyperspectral camouflaged object detection.
The core contributions of this work are summarized as follows:
\begin{itemize}[leftmargin=*, nosep]
    \item We introduce HyperCOD, the first large-scale and challenging dataset for hyperspectral camouflaged object detection, covering diverse scenes and multiple camouflage scenarios.
    \item We propose HSC-SAM, a unified spectral-aware detection framework that integrates spectral-spatial decomposition and complementary prompt learning to adapt SAM to the hyperspectral domain.
    \item We design Spectral-Guided Token Dropout (SGTD) and Fusion Detail Enhancer (FDE) modules to improve saliency-guided pruning and boundary refinement, resulting in precise segmentation performance.
\end{itemize}

\section{Related Work}
\paragraph{Camouflaged Object Detection Datasets.}  
RGB-based camouflaged object detection has advanced with datasets like CHAMELEON~\shortcite{paper41}, CAMO-V1~\shortcite{paper42}, COD10K~\shortcite{paper52}, and NC4K~\shortcite{paper44}. However, their limited spectral resolution struggles with foreground-background ambiguity. For hyperspectral analysis, HS-SOD~\shortcite{paper32} was the first dataset tailored for HSOD, though limited to 60 images. HSOD-BIT~\shortcite{paper28} and its recent version HSOD-BIT-V2~\shortcite{paper33} expanded the scale to 319 annotated images and introduced various challenges such as non-uniform illumination and overexposure. As these focus on saliency detection, we introduce \textbf{HyperCOD}, the first large-scale hyperspectral COD dataset featuring diverse real-world camouflage challenges, establishing a robust benchmark for advancing HCOD research.

\paragraph{Camouflaged Object Detection Methods.}  
Traditional COD methods~\shortcite{paper46,paper47,paper48} rely on hand-crafted features (\textit{e.g.}, texture, color, motion), which struggle with complex backgrounds and variable object appearances, leading to poor robustness. Recent models (SINet~\shortcite{paper3}, D2CNet~\shortcite{paper10}) employ attention mechanisms, while edge-guided approaches (SegMaR~\shortcite{paper14}, BGNet~\shortcite{paper15}) integrate boundary cues. Other methods, such as ZoomNet~\shortcite{paper3} and FSEL~\shortcite{paper18}, address modality-specific challenges via confidence-aware learning and feature fusion. Despite their success, these methods remain confined to RGB modality. Our framework overcomes this by fully leveraging hyperspectral information.

\paragraph{SAM-based Methods.}  
While SAM adaptations like SAM-COD~\shortcite{paper19} and SAM-Adapter~\shortcite{paper20} have advanced COD through efficient tuning strategies, and approaches such as ProMAC~\shortcite{paper21} and MD-SAM~\shortcite{paper22} have explored multi-scale processing, these methods remain constrained to RGB inputs and manual prompting. Our framework addresses both limitations by automatically extracting discriminative spectral features while eliminating the need for handcrafted prompts.

\section{HyperCOD dataset}

Our HyperCOD dataset comprises 350 high-quality hyperspectral images, each containing 200 spectral bands spanning 400–1000 nm with a spatial resolution of 1240 $\times$ 1680 pixels. The dataset is partitioned into a training set (280 samples) and a testing set (70 samples) at a 4:1 ratio.

\begin{table}[t] \footnotesize
\centering
\setlength{\tabcolsep}{0.2pt}
\begin{tabular}{l|cccc}
\toprule
\textbf{Dataset} & \textbf{Modality} & \textbf{Resolution} & \textbf{Channels} \\
\midrule
CHAMELEON~\shortcite{paper41} & RGB     & 450$\times$300$\sim$2304$\times$3456 & 3        \\
CAMO-V1~\shortcite{paper42}   & RGB     & 154$\times$156$\sim$7360$\times$4912 & 3       \\
CAMO++~\shortcite{paper51}    & RGB     & Unreleased                           & 3         \\
COD10K~\shortcite{paper52}    & RGB     & 300$\times$199$\sim$2976$\times$3968 & 3         \\
NC4K~\shortcite{paper44}      & RGB     & 354$\times$268$\sim$1280$\times$960  & 3        \\
ACOD-12K~\shortcite{paper54}  & RGB-D   & Unreleased                           & 4  \\
\midrule
\textbf{HyperCOD} (Ours)  & \textbf{HSI} & 1680$\times$1240 & \textbf{200} \\
\bottomrule
\end{tabular}%
\caption{Comparison of Existing COD Datasets. \textbf{HyperCOD} is the first dataset to introduce HSIs for COD.}
\label{tab:cod_datasets}
\end{table}

\begin{figure}[t]
\centering
\includegraphics[width=\linewidth]{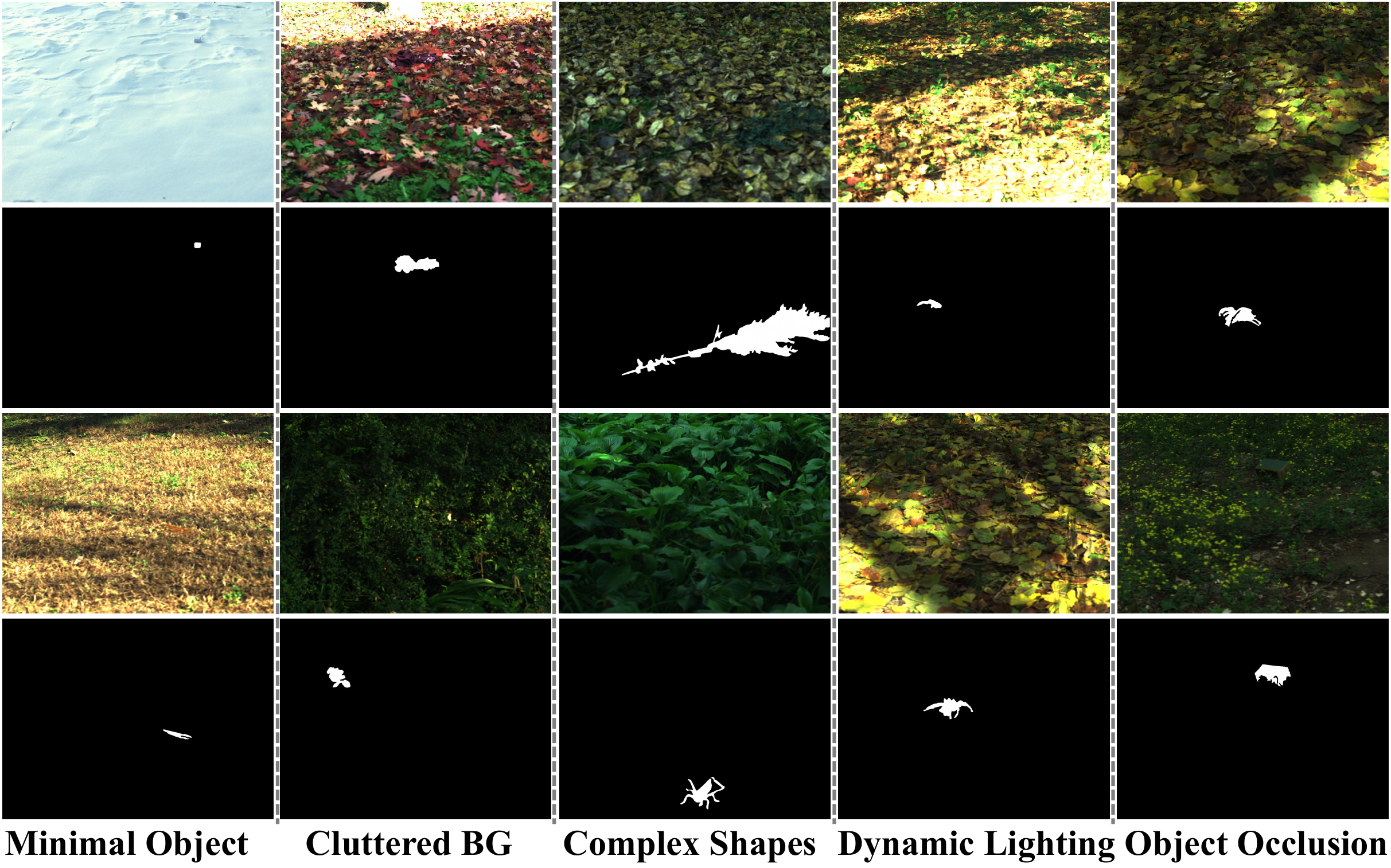}
\caption{Representative samples and corresponding ground truth masks from five challenging scenarios.}
\label{fig:scene_examples}
\end{figure}

\paragraph{Comparison with Existing Datasets.}  
As shown in Tab.~\ref{tab:cod_datasets}, HyperCOD surpasses existing camouflaged object detection and hyperspectral object detection datasets in several key aspects. While RGB-based COD datasets are well-established, their single-modality nature restricts the utilization of spectral information. HyperCOD is the first dataset tailored for hyperspectral COD, offering high-resolution spatial details and dense spectral coverage. It ensures greater data diversity and introduces more complex and challenging scenarios for the evaluation of robust hyperspectral camouflaged object detection models.

\paragraph{Data Collection.}  
HyperCOD was constructed using the HySpex Mjolnir-1024 hyperspectral camera coupled with a STANDA 8 MR 190 precision motorized rotation stage for line-by-line scanning of natural scenes. To ensure diversity, data were captured under varying weather conditions (e.g., cloudy, sunny) and times of day (morning, afternoon), while maintaining consistent imaging parameters such as exposure time, white balance, and gain compensation.

\paragraph{Annotation Process.}  
Pixel-level ground truth annotations were created on false-color images using MATLAB's \texttt{ImageLabeler} toolbox. Camouflaged objects were labeled as foreground (1), while the remaining regions were marked as background (0). Each sample in the dataset includes a hyperspectral image (.mat), a binary segmentation mask (.png), and a pseudo-RGB image (.jpg).

\paragraph{Challenge Attributes.}
HyperCOD introduces significant challenges categorized into five key attributes: (i) Minimal Objects (MO) featuring extremely small and inconspicuous targets, (ii) Complex Shapes (CS) with irregular or fragmented boundaries, (iii) Dynamic Lighting (DL) under strong illumination variations, (iv) Object Occlusion (OO) with partially obscured objects, and (v) Cluttered Backgrounds (CB) in environmentally complex scenes---as depicted in Fig.~\ref{fig:scene_examples}. These attributes collectively highlight the critical role of spectral information in camouflage detection, imposing higher demands on detection models while providing a rigorous benchmark for hyperspectral frameworks.

\paragraph{Statistical Analysis.}

We perform a comprehensive statistical analysis of HyperCOD across multiple dimensions to ensure diversity, balance, and challenge.

\begin{figure}[t]
\centering
\includegraphics[width=\linewidth]{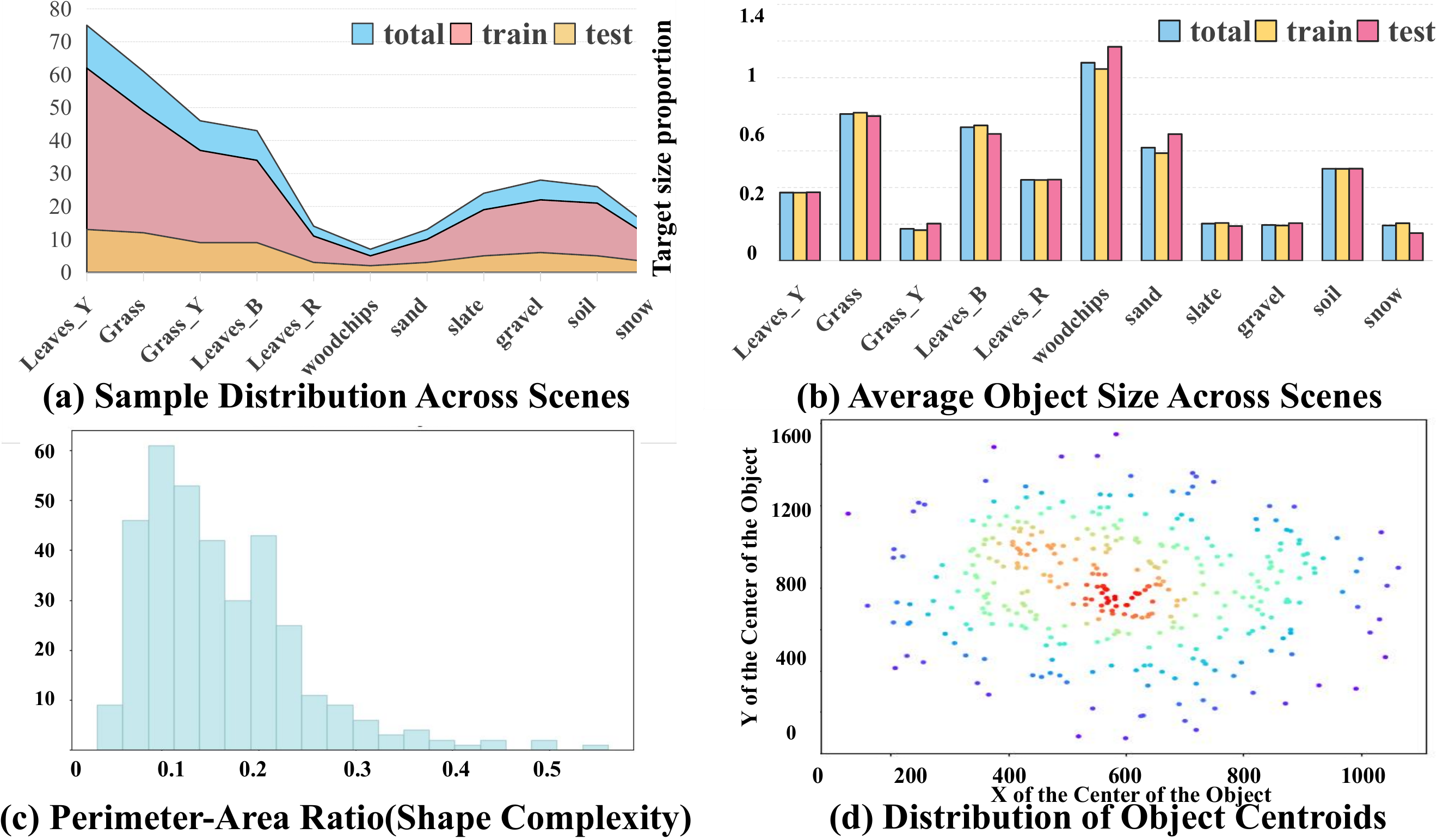}
\caption{Statistics of the HyperCOD Dataset.}
\label{fig:object_stats}
\end{figure}

\noindent\textbf{\textit{Scene diversity.}} is ensured by including a wide range of natural environments. The training and testing splits are carefully balanced across scene categories, as shown in Fig.~\ref{fig:object_stats}(a).

\noindent\textbf{\textit{Object size distribution.}} is analyzed in Fig.~\ref{fig:object_stats}(b). Most targets in HyperCOD are relatively small, increasing the dataset's difficulty. Samples with object-to-image area ratios below 0.5\% are classified as \emph{tiny-object cases}, with proportional representation in both training and testing subsets to maintain consistent difficulty.

\noindent\textbf{\textit{Boundary complexity.}} is evaluated using the edge-to-perimeter ratio (Fig.~\ref{fig:object_stats}(c)). Targets with ratios exceeding 0.3 are labeled as \emph{complex-edge cases}, requiring robust segmentation and fine-grained contour learning. These examples encourage models to better capture subtle structural details.

\noindent\textbf{\textit{Spatial distribution.}} is examined in Fig.~\ref{fig:object_stats}(d), which shows the centroid distribution of target objects. HyperCOD exhibits minimal center bias, ensuring a uniform spatial distribution. This property mitigates common pitfalls in existing datasets and promotes fairer evaluation, encouraging models to generalize across diverse object positions.

\begin{figure*}[tp]
    \centering
    \includegraphics[width=\textwidth]{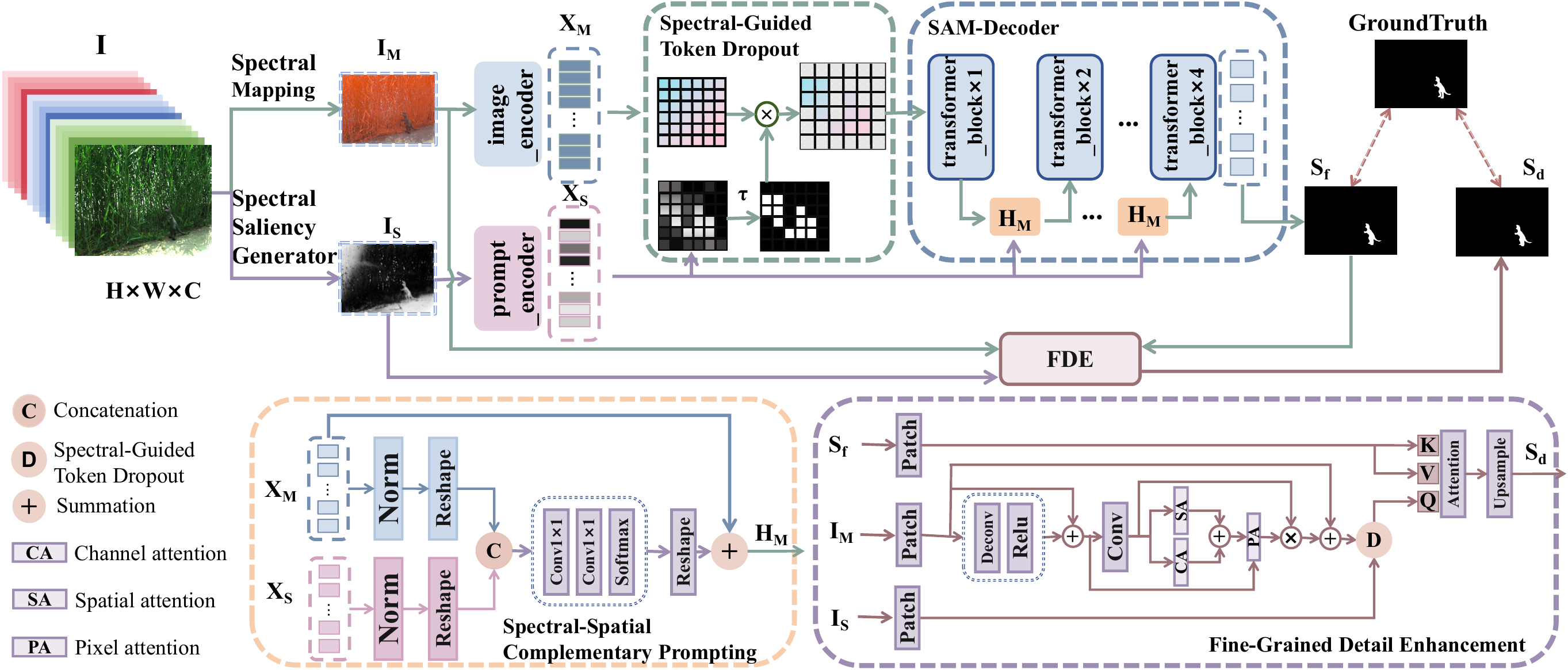}
    \caption{Overview of the proposed HSC-SAM framework. Spatial features $\boldsymbol{I}_M$ are used in the SAM image encoder, while spectral saliency $\boldsymbol{I}_S$ guides prompt encoding and token pruning. Joint spatial-spectral learning enhances object localization, and a Fusion Detail Enhancer (FDE) refines boundary details for accurate segmentation.}
    \label{fig:framework}
\end{figure*}

\section{Method}

The goal of the hyperspectral camouflaged object detection task is to localize camouflaged objects from a hyperspectral image cube \( \boldsymbol{I} \in \mathbb{R}^{\mathrm{H} \times \mathrm{W} \times \mathrm{C}} \), where \( \mathrm H \), \( \mathrm W \), and \(\mathrm C \) denote the spatial height, width, and spectral channels, respectively, and to generate a corresponding saliency map \( \boldsymbol{F} \in \mathbb{R}^{\mathrm H \times \mathrm W \times 1} \) that highlights object regions. To address this task, we propose the HyperSpectral Camouflage-aware SAM (HSC-SAM) framework, as illustrated in Fig.~\ref{fig:framework}. 

HSC-SAM employs a Spectral-Spatial Decomposition Module (SSDM) to extract complementary spatial and spectral features from hyperspectral imagery. The spatial map preserves structural details for the SAM image encoder, while the spectral saliency map highlights material differences for the prompt encoder. This spectral prompt further guides token pruning to suppress non-salient background activations. During decoding, spatial-spectral joint learning enhances object localization to produce initial segmentations. Finally, the Fusion Detail Enhancer (FDE) module supervises high-frequency boundary restoration during training, yielding more precise segmentation results.

\subsection{Spectral-Spatial Decomposition Module}
\paragraph{Color Space Mapping.}
Metamerism causes spectrally distinct materials to appear identical in RGB images due to sensor limitations, posing challenges for camouflaged object detection. In contrast, the CIE XYZ color space integrates spectral reflectance with standardized color matching functions across the visible spectrum, preserving material-dependent spectral nuances.

To construct the transformed image $\boldsymbol{I}_M$, we uniformly sample 33 representative bands from the original 200 spectral channels and apply the standard CIE color matching functions $\mathrm{W}_t$ ~\shortcite{foster2004} to map them into the CIEXYZ color space. For a hyperspectral image $\boldsymbol{I}(\lambda)$, the tristimulus values $(X, Y, Z)$ are computed as:
\begin{equation}
    \boldsymbol{I}_M = \sum_{i=1}^{\mathrm{N}} \boldsymbol{I}(\lambda_i) \cdot \mathrm{W}_t(\lambda_i), \quad t \in \{X, Y, Z\},
\end{equation}
where $\boldsymbol{I}(\lambda_i)$ is the spectral intensity at wavelength $\lambda_i$, $\mathrm{W}_t(\lambda_i)$ is the color matching function for channel $t$, and $\mathrm{N}$ is the number of spectral bands. This preserves material-dependent spectral signatures that RGB sensors lose.

\paragraph{Spectral Saliency Generator.}
While the aforementioned steps enhance the discriminability of foreground and background in terms of spatial structure, they do not fully exploit spectrum. To address this limitation, we employ SSG proposed by Liu~\textit{et al.}~\shortcite{paper27} to generate a spectral prompt $\boldsymbol{I}_S \in \mathbb{R}^{\mathrm H \times \mathrm W \times 3}$, which encodes spectral saliency cues. It constructs an $N$-level Gaussian pyramid from the input hyperspectral image and computes spectral angular distances (SAD) between feature vectors extracted from non-adjacent pyramid levels. For a pixel $(i,j)$, let $v_c^{(i,j)}$ and $v_{c+3}^{(i,j)}$ represent spectral vectors from levels $c$ and $c+3$, respectively. The intermediate saliency map at level $c$ is defined as:

\begin{equation}
\boldsymbol{S}^{(c)}(i,j) = \arccos \left( \frac{v_c^{(i,j)} \cdot v_{c+3}^{(i,j)}}{\|v_c^{(i,j)}\|_2 \cdot \|v_{c+3}^{(i,j)}\|_2} \right).
\end{equation}

The spectral prompt $\boldsymbol{I}_S$ is then obtained by concatenating the intermediate saliency maps across levels $c\in\{2,3,4\}$.

In our framework, $\boldsymbol{I}_M$ is used as the encoder input, while $\boldsymbol{I}_S$ serves as a spectral prompt to guide localization.

\subsection{Spectral-Guided Token Dropout}

To suppress background noise and reduce computational overhead, we introduce a Spectral-Guided Token Dropout (SGTD) mechanism based on the spectral saliency map. This mechanism prunes background tokens at the token level by leveraging spectral attention, thus improving model focus on salient regions and enhancing feature discrimination.

Given the encoded image features $\mathbf{\boldsymbol{X}_\mathrm{M}} \in \mathbb{R}^\mathrm{B \times N \times C}$ and the corresponding patch-wise spectral saliency map $\mathbf{\boldsymbol{X}}_\mathrm{S} \in \mathbb{R}^\mathrm{B \times N \times C}$, we first compute the mean activation across the channel dimension to obtain a saliency score for each token:
\begin{equation}
\mathbf{\boldsymbol{S}}_\mathrm{score} = \frac{1}{\mathrm C} \sum_{c=1}^{\mathrm C} \mathbf{\boldsymbol{X}}_\mathrm{S}^{(:, :, c)} \in \mathbb{R}^\mathrm{B \times N \times 1},
\end{equation}
where $\mathbf{S}_\mathrm{score}$ denotes the scalar saliency response per token.

We then apply a user-defined threshold $\tau$ to binarize the saliency map and generate a token-wise dropout mask:
\begin{equation}
\mathbf{\boldsymbol{M}}_\mathrm{mask}^{(i,j)} = 
\begin{cases}
1, & \text{if } \mathbf{S}_\mathrm{score}^{(i,j)} \geq \tau, \\
0, & \text{otherwise},
\end{cases}
\end{equation}
where $i$ indexes the batch and $j$ indexes the token position.

The binary mask is finally applied to the encoded image features to suppress background information:
\begin{equation}
\widetilde{\mathbf{\boldsymbol{X}}}_\mathrm{M} = \mathbf{\boldsymbol{X}}_\mathrm{\boldsymbol{M}} \odot \mathbf{M}_\mathrm{mask},
\end{equation}
where $\odot$ denotes element-wise multiplication with broadcasting along the channel dimension.

This dropout operation directly integrates spectral priors into token selection, enabling the model to dynamically focus on high-saliency regions while discarding low-importance tokens, thus improving both detection precision and computational efficiency.

\subsection{Spectral-Spatial Complementary Prompting}
To align spectral saliency with spatial structural cues for more accurate target localization, we introduce a Spectral-Spatial Complementary Prompting mechanism. The hyperspectral branch provides rich spectral priors, enabling the model to identify subtle material variations that are imperceptible in RGB imagery, especially under challenging camouflage.

Given image tokens \(\mathbf{\boldsymbol{X}}_\mathrm{M}^{i} \in \mathbb{R}^\mathrm{B \times L \times C}\) and spectral tokens \(\mathbf{\boldsymbol{X}}_\text{S} \in \mathbb{R}^\mathrm{B \times L \times C}\), we first apply layer normalization:
\begin{equation}
\tilde{\mathbf{\boldsymbol{H}}}_\mathrm{M}^{i} = \text{LN}(\mathbf{\boldsymbol{X}}_\mathrm{M}^{i}), \quad \tilde{\mathbf{\boldsymbol{H}}}_\text{S} = \text{LN}(\mathbf{\boldsymbol{X}}_\text{S}) .
\end{equation}

Each normalized token sequence is reshaped into 2D feature maps using a token-to-feature transformation \(\phi: \mathbb{R}^\mathrm{B \times L \times C} \rightarrow \mathbb{R}^\mathrm{B \times C \times H \times W}\):
\begin{equation}
\mathbf{\boldsymbol{F}}_\text{img} = \phi(\tilde{\mathbf{\boldsymbol{H}}}_\mathrm{M}^{i}), \quad \mathbf{\boldsymbol{F}}_\text{spec} = \phi(\tilde{\mathbf{\boldsymbol{H}}}_\text{S}) .
\end{equation}

The two features are concatenated along the channel dimension and fused via a prompt block \(\mathcal{P}(\cdot)\), which integrates spectral saliency and spatial structure:
\begin{equation}
\mathbf{\boldsymbol{F}}_\text{fused} = \mathcal{P} \left( \text{Concat}(\mathbf{\boldsymbol{F}}_\text{img}, \mathbf{\boldsymbol{F}}_\text{spec}) \right) .
\end{equation}

The fused features are projected back to token space and added to the original image tokens in a residual manner:
\begin{equation}
\hat{\mathbf{\boldsymbol{H}}}_\mathrm{M}^{i+1} = \mathbf{\boldsymbol{H}}_\mathrm{M}^{i} + \phi^{-1}(\mathbf{\boldsymbol{F}}_\text{fused}) .
\end{equation}

This complementary prompting mechanism allows the model to refine spatial attention using spectral contrast, promoting fine-grained camouflage object segmentation even in low-contrast or texture-similar environments.

\subsection{Fine-Grained Detail Enhancement}

To recover boundary details lost in high-level semantic decoding, we introduce a Fine-Grained Detail Enhancement (FDE) module that injects low-level spatial features via joint attention modulation.

Let the spectral-mapped pseudo-RGB input be \( \boldsymbol{I}_M \in \mathbb{R}^\mathrm{B \times C \times H \times W} \). We extract structural features using depthwise separable convolution:
\begin{equation}
\boldsymbol{F} = \phi(\boldsymbol{I}_M), \quad \boldsymbol{F} \in \mathbb{R}^\mathrm{B \times C \times H \times W}.
\end{equation}

To refine \( \boldsymbol{F} \), we apply three types of attention. Channel Attention (CA) computes inter-channel dependencies:
\begin{equation}
\boldsymbol{A}_\text{CA} = \sigma \left( W_2 \delta \left( W_1 \cdot \text{GAP}(\boldsymbol{F}) \right) \right) \in \mathbb{R}^\mathrm{B \times C \times 1 \times 1},
\end{equation}
where \( \text{GAP} \) denotes global average pooling, \( \delta \) is ReLU, and \( \sigma \) is sigmoid, \( W_1 \) and \( W_2 \) are convolutional layers whose weights are learned during training. Spatial Attention (SA) captures boundary-localized responses:
\begin{equation} \small
\boldsymbol{A}_{\text{SA}} = \sigma\Bigl( \text{Conv}_{k\times k} \bigl( \text{AvgPool}_c(\boldsymbol{F}) \oplus \text{MaxPool}_c(\boldsymbol{F}) \bigr) \Bigr),
\end{equation}
where $\oplus$ denotes channel-wise concatenation, $\text{Conv}_{k \times k}$ is a convolution kernel (e.g., $k=7$), and $\boldsymbol{A}_{\text{SA}} \in \mathbb{R}^{B \times 1 \times H \times W}$.
\begin{equation}
\boldsymbol{A}_\text{PA} = \sigma(\text{Conv}_{1\times1}(\boldsymbol{F})), \quad \boldsymbol{A}_\text{PA} \in \mathbb{R}^\mathrm{B \times C \times H \times W}.
\end{equation}

The overall modulation map is computed as:
\begin{equation}
\boldsymbol{A} = \boldsymbol{A}_\text{CA} \cdot \boldsymbol{A}_\text{SA} \cdot \boldsymbol{A}_\text{PA}, \quad \boldsymbol{A} \in \mathbb{R}^\mathrm{B \times C \times H \times W}.
\end{equation}
We then enhance the decoder prediction \( \boldsymbol{S}_f \in \mathbb{R}^\mathrm{B \times 1 \times H \times W} \) with the modulated fine-grained features:
\begin{equation}
\boldsymbol{S}_d = \boldsymbol{S}_f + \boldsymbol{A} \odot \boldsymbol{F},
\end{equation}
where \( \odot \) denotes element-wise multiplication.

This formulation integrates boundary-aware low-level features into the segmentation pipeline, yielding sharper and more complete object delineation.

\begin{table}[!t]
\centering \small
\setlength{\tabcolsep}{2pt}
\begin{tabular}{l|cccc|cc}
\toprule
\textbf{Method} & \textbf{MAE} $\downarrow$ & \textbf{E} $\uparrow$ & \textbf{S} $\uparrow$ & \textbf{Adp-F} $\uparrow$ & \textbf{Params} & \textbf{FLOPs} \\
\midrule
\multicolumn{7}{c}{\textbf{\textit{RGB-based Methods}}} \\
\midrule
SINet-V2    & 0.0033 & 0.732  & 0.746  & 0.480 & 27.0    & 12.2    \\
ZoomNet     & 0.0044 & 0.831  & 0.757 & 0.352 & 32.4    & 203.5    \\
FRINet     & 0.0027 & 0.889 & 0.759  & 0.604 & 100.0      & 112.7    \\
SAM2-UNet  & 0.0022 & \textbf{0.899} & \textbf{0.805} & 0.641 & 216.4   & 128.4    \\
HGINet    & 0.0039 & 0.850 & 0.766  & 0.585 & 400.0      & 530.4    \\
Camoformer   & 0.0070 & 0.673 & 0.355 & 0.660  & 71.4    & 41.8     \\
\midrule
\multicolumn{7}{c}{\textbf{\textit{Hyperspectral Salient Object Detection Methods}}} \\
\midrule
SAD    & 0.1505 & 0.325  & 0.483  & 0.0061 & --        & --        \\
DMSSN      & 0.0295 & 0.687  & 0.446  & 0.409 & 31.0       & \textbf{8.1}     \\
SMN-PVT    & 0.0066 & 0.654  & 0.531  & 0.103 & \textbf{10.2}    & 14.8     \\
Hyper-HRNet     & 0.0149 & 0.818  & 0.608  & 0.150 & 29.6    & 19.0       \\
\midrule
\multicolumn{7}{c}{\textbf{\textit{Hyperspectral Camouflage Object Detection Methods}}} \\
\midrule
\textbf{HSC-SAM}  & \textbf{0.0017} & 0.853 & 0.802 & \textbf{0.681} & 11.7 & 94.2  \\
\bottomrule
\end{tabular}
\caption{Results on the HyperCOD dataset. Bold values indicate the best performance in each column.}
\label{tab:hypercod_comparison}
\end{table}

\subsection{Learning Objective}
We adopt a composite loss function that combines Binary Cross-Entropy (BCE) and Intersection-over-Union (IoU) to supervise both the decoder output and the final prediction. This encourages accuracy and structural consistency.

The loss for the decoder output \( \boldsymbol{S}_d \) is defined as:
\begin{equation}
    \mathcal{L}_{\text{dec}} = \mathcal{L}_{\text{BCE}}(\boldsymbol{S}_d, \boldsymbol{G}) + \mathcal{L}_{\text{IoU}}(\boldsymbol{S}_d, \boldsymbol{G}),
\end{equation}
and the loss for the final prediction \( \boldsymbol{S}_f \) as:
\begin{equation}
    \mathcal{L}_{\text{final}} = \mathcal{L}_{\text{BCE}}(\boldsymbol{S}_f, \boldsymbol{G}) + \mathcal{L}_{\text{IoU}}(\boldsymbol{S}_f, \boldsymbol{G}).
\end{equation}

The total loss is then given by \( \mathcal{L}_{\text{total}} = \mathcal{L}_{\text{dec}} + \mathcal{L}_{\text{final}} \).

\section{Experiment}

\begin{table*}[!t]
\centering \small
\setlength{\tabcolsep}{5pt}
\begin{tabular}{l|cc|cc|cc|cc|cc} 
\toprule
\multirow{2}{*}{\textbf{Method}} & \multicolumn{2}{c|}{\textbf{Cluttered Backgrounds}} & \multicolumn{2}{c|}{\textbf{Complex Shapes}} & \multicolumn{2}{c|}{\textbf{Dynamic Lighting}} & \multicolumn{2}{c|}{\textbf{Multiple Objects}} & \multicolumn{2}{c}{\textbf{Occlusion}}  \\ 
\cmidrule{2-11}
                                 & MAE~$\downarrow$ & Adp-F~$\uparrow$                 & MAE~$\downarrow$ & Adp-F~$\uparrow$          & MAE~$\downarrow$ & Adp-F~$\uparrow$            & MAE~$\downarrow$ & Adp-F~$\uparrow$            & MAE~$\downarrow$ & Adp-F~$\uparrow$     \\ 
\midrule
DMSSN~\shortcite{paper28}             & 0.0762           & 0.0737                           & 0.0809           & 0.0793                    & 0.0784           & 0.0110                      & 0.0818           & 0.0096                      & 0.1223           & 0.0544               \\
Hyper-HRNet~\shortcite{paper33}       & 0.0157           & 0.2243                           & 0.0161           & 0.2025                    & 0.0137           & 0.1581                      & 0.0151           & 0.0383                      & 0.0095           & 0.3183               \\
SMN~\shortcite{paper27}               & 0.0105           & 0.0826                           & 0.0103           & 0.1087                    & 0.0068           & 0.0836                      & 0.0036           & 0.0344                      & 0.0081           & 0.2416               \\
HGINet~\shortcite{paper58}            & 0.0049           & 0.7451                           & 0.0061           & 0.6635                    & 0.0034           & 0.7277                      & 0.0017           & 0.5176                      & 0.0020           & 0.8486               \\
SAM2  -UNet~\shortcite{paper59}          & 0.0036           & 0.6613                           & 0.0029           & 0.6712                    & 0.0022           & 0.7219                      & \textbf{0.0010}  & 0.5600                      & 0.0022           & 0.8261               \\
SINet-V2~\shortcite{paper52}          & 0.0054           & 0.5937                           & 0.0058           & 0.5322                    & 0.0036           & 0.4945                      & 0.0016           & 0.3721                      & 0.0028           & 0.7941               \\
ZoomNet~\shortcite{paper3}            & 0.0073           & 0.3726                           & 0.0064           & 0.4511                    & 0.0050           & 0.4135                      & 0.0021           & 0.2028                      & 0.0041           & 0.6220               \\ 
\midrule
\textbf{HSC-SAM} (Ours)          & \textbf{0.0033}  & \textbf{0.7588}                  & \textbf{0.0026}  & \textbf{0.7727}           & \textbf{0.0015}  & \textbf{0.8242}             & 0.0011           & \textbf{0.5828}             & \textbf{0.0012}  & \textbf{0.8698}      \\
\bottomrule
\end{tabular}
\caption{Performance comparison under five challenging attributes.}
\label{tab:attribute_eval}
\vspace{-2mm}
\end{table*}

\begin{figure*}[!t]
    \centering
    \includegraphics[width=0.95\linewidth]{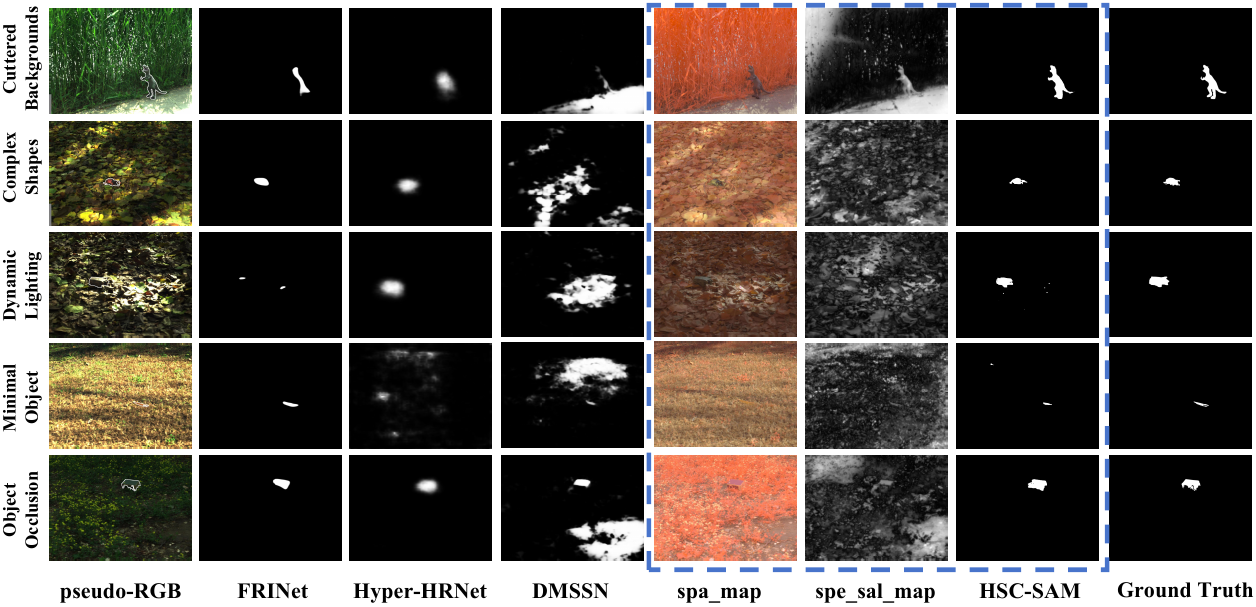}
    \caption{Qualitative results on HyperCOD. HSC-SAM offers clearer contours in challenging camouflage scenarios.}
    \label{fig:qualitative_comparison}
    \vspace{-2mm}
\end{figure*}

\subsection{Results on HyperCOD}
\paragraph{Quantitative Results.}
We compare HSC-SAM with representative RGB-based camouflaged object detection models on the HyperCOD dataset, as well as several state-of-the-art hyperspectral salient object detection methods for reference. As shown in Tab.~\ref{tab:hypercod_comparison}, RGB-based models built upon SAM’s backbone generally outperform conventional camouflaged object detection models, benefiting from their strong generalization capability. Building on this foundation, our HSC-SAM further bridges the performance gap by effectively leveraging spectral cues, achieving state-of-the-art results across all key metrics. Specifically, HSC-SAM attains the lowest MAE of \textbf{0.0017}, outperforming the second-best result by 0.0005, and achieves the highest Adp-F score of \textbf{0.681}, representing a 3.1\% improvement over the closest competitor, Camoformer. These results demonstrate that incorporating hyperspectral information enables more accurate object localization and finer contour delineation beyond what can be achieved with RGB-based SAM variants alone.

\paragraph{Efficiency Analysis.}
As shown in Tab.~\ref{tab:hypercod_comparison}, HSC-SAM achieves competitive complexity with \textbf{11.7M} parameters and \textbf{94.2G} FLOPs, which are substantially lower than those of other SAM-based variants such as SAM2-UNet (216.4M, 128.4G) and high-capacity RGB models like HGINet. These results highlight the effectiveness of HSC-SAM in balancing accuracy and computational efficiency for hyperspectral camouflaged object detection.

\paragraph{Attribute-based Comparison.}

HSC-SAM demonstrates superior robustness across all challenging scenarios (Tab.~\ref{tab:attribute_eval}). The Spectral-Spatial Decomposition Module effectively handles dynamic lighting conditions by revealing spectrally-distinct features, while spectral saliency suppresses background clutter through adaptive token pruning. The framework further refines localization via Spectral-Spatial Complementary Prompting and enhances boundary precision through Fine-Grained Detail Enhancement during training.

\paragraph{Qualitative Results.}
Fig.~\ref{fig:qualitative_comparison} illustrates the prediction results of our HSC-SAM compared with other camouflaged object detection methods. In complex and challenging scenarios, HSC-SAM accurately localizes spatially camouflaged yet spectrally salient objects and precisely segments their shapes. Compared to other approaches, HSC-SAM delivers segmentation results with notably sharper and more accurate object boundaries.

\begin{figure}[!t]
    \centering
    \includegraphics[width=\linewidth]{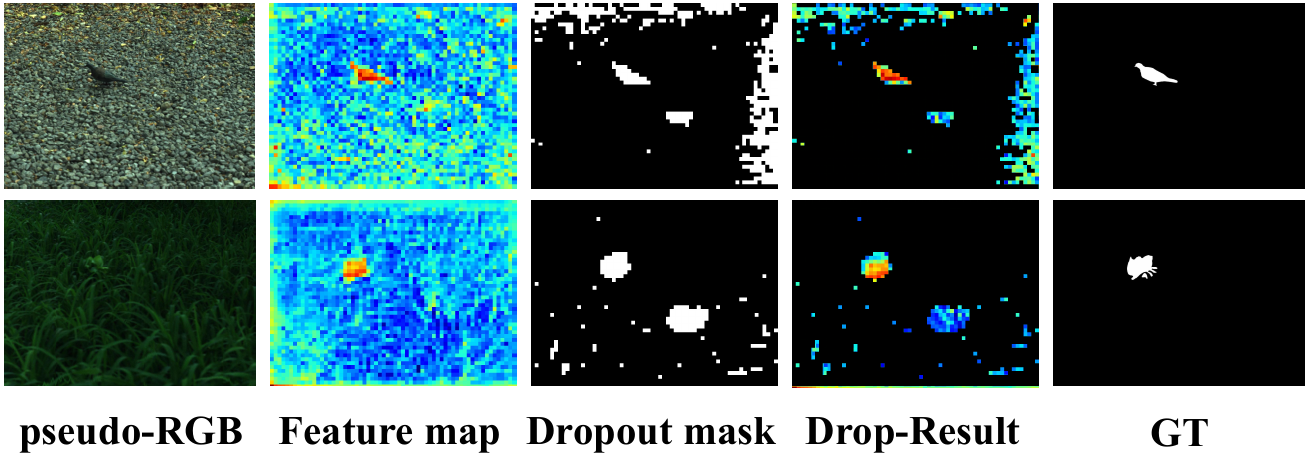}
    \caption{Visualization of the SGTD feature map. SGTD filters low-saliency background features using spectral cues, enabling precise early-stage localization.}
    \label{fig:sgtd}
\end{figure}

\paragraph{Visualization of Background Dropout.}
To intuitively demonstrate the superiority of HSC-SAM, we visualize its intermediate behavior for qualitative analysis. Fig.~\ref{fig:sgtd} illustrates how SGTD leverages the spectral saliency map to suppress background features. By exploiting spectral cues, HSC-SAM effectively highlights target regions and filters out background noise at an early stage, significantly reducing computational overhead.

\begin{table}[!t]
\centering \small
\setlength{\tabcolsep}{4pt}
\begin{tabular}{ccccc|cc}
\toprule
Light-SAM & SSDM & SSCP & SGTD & FDE & Adp-F $\uparrow$ & E $\uparrow$ \\
\midrule
\ding{51} & \ding{72}  & \ding{55}    & \ding{55}    & \ding{55}   & 0.481 & 0.782 \\
\ding{51} & \ding{51}  & \ding{55}    & \ding{55}    & \ding{55}   & 0.557 & 0.806 \\
\ding{51} & \ding{51}  & \ding{51}    & \ding{55}    & \ding{55}   & 0.596 & 0.823 \\
\ding{51} & \ding{51}  & \ding{51}    & \ding{51}    & \ding{55}   & 0.637 & 0.831 \\
\ding{51} & \ding{51}  & \ding{51}    & \ding{51}    & \ding{51}   & 0.657 & 0.838 \\
\ding{51} & \ding{51}  & \ding{51}    & \ding{51}    & \ding{169}  & \textbf{0.681} & \textbf{0.853} \\
\bottomrule
\end{tabular}
\caption{Ablation study of key components in HSC-SAM. \ding{72}: pseudo-RGB input only; \ding{169}: training-only module.}
\label{tab:ablation}
\end{table}

\subsection{Ablation Study} 
\begin{figure}[!t]
    \centering
    \includegraphics[width=\linewidth]{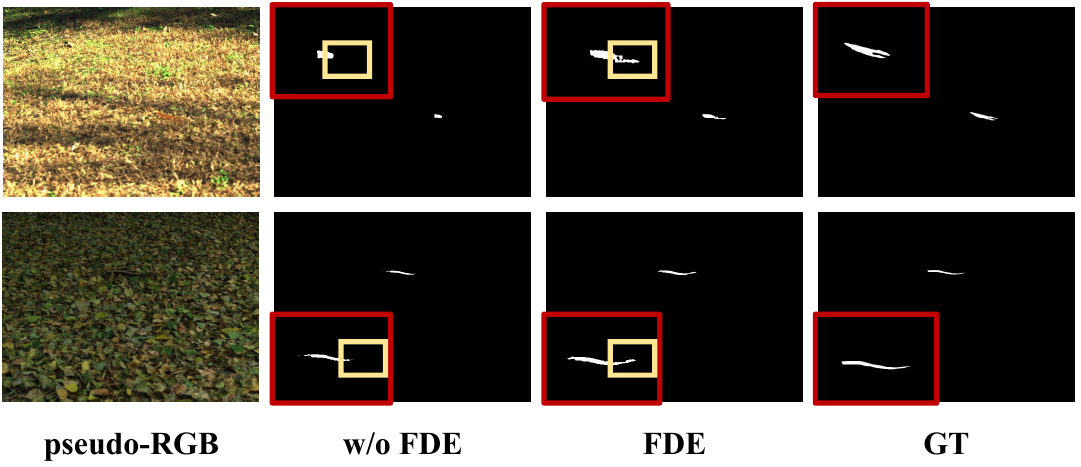}
    \caption{Visual comparison of predictions with and without training-time FDE supervision. FDE-enhanced models recover finer structural details and more accurate boundaries.}
    \label{fig:fde_effect}
\end{figure}

\paragraph{Effect of Key Components.}
We conduct an ablation study to assess the contribution of each component in our HSC-SAM framework, as shown in Tab.~\ref{tab:ablation}. The Spectral-Spatial Decomposition Module replaces pseudo-RGB inputs with CIEXYZ-mapped images, achieving a \textbf{15.6\%} Adp-F improvement by preserving material-specific spectral signatures that are critical for distinguishing true camouflage. Spectral-Spatial Complementary Prompting further increases Adp-F to 0.596 (\textbf{7.0\%} gain) and E-measure to 0.823 (2.1\% gain) by establishing joint spectral-spatial reasoning paths. The Spectral-Guided Token Dropout achieves \textbf{6.9\%} Adp-F improvement and 1.1\% E-measure gain through adaptive background suppression. This highlights the module's effectiveness. Finally, the Fine-Grained Detail Enhancement module enhances boundary precision through training-time multi-scale gradient learning, leaving a \textbf{6.9\%} Adp-F legacy and producing sharper contours (Fig.~\ref{fig:fde_effect}) even when disabled during inference.

\begin{table}[!t]
\centering \small
\setlength{\tabcolsep}{12pt}
\begin{tabular}{c|cccc}
\toprule
$\tau$ & MAE $\downarrow$ & Adp-F $\uparrow$ & E $\uparrow$ & S $\uparrow$ \\
\midrule
0.1    & 0.0018          & 0.668            & 0.796        & 0.783        \\
0.03   & 0.0020          & 0.663            & 0.861        & 0.788        \\
\textbf{0.01}   & \textbf{0.0017}          & \textbf{0.680}            & \textbf{0.853}        & \textbf{0.802}        \\
0.003  & 0.0020          & 0.680            & 0.832        & 0.789        \\
0.001  & 0.0019          & 0.659            & 0.795        & 0.782        \\
\bottomrule
\end{tabular}
\caption{Ablation study on the Spectral-Guided Token Dropout (SGTD) threshold $\tau$.}
\label{tab:sgtd_threshold}
\end{table}

\begin{table}[!t]
    \centering \small
    \setlength{\tabcolsep}{9pt}
    \begin{tabular}{lccc}
        \toprule
        \textbf{Setting} & \textbf{Parameters} & \textbf{FPS} & \textbf{Adp-F $\uparrow$} \\
        \midrule
        w/ SGTD    & 12.09M           & 19.1         & 0.639 \\
        w/o SGTD   & 11.74M           & 20.3 (6.3\% $\uparrow$) & 0.681 \\
        \bottomrule
    \end{tabular}
    \caption{Efficiency analysis of the SGTD module.}
    \label{tab:sgtd_efficiency}
\end{table}

\paragraph{SGTD Module Analysis.}
We present a comprehensive analysis of the SGTD module, covering both its efficiency and the sensitivity of its key hyperparameter.
The efficiency analysis (Table~\ref{tab:sgtd_efficiency}) demonstrates a favorable trade-off: despite a marginal increase in parameter count and a minor reduction in FPS, the proposed module yields a substantial gain in detection accuracy.
Furthermore, we evaluate the robustness of the SGTD threshold $\tau$, which binarizes spectral saliency maps to guide the token dropout. As shown in Tab.~\ref{tab:sgtd_threshold}, $\tau=0.01$ achieves the optimal balance across all metrics, confirming the module's stability to minor hyperparameter variations.

\section{Conclusion}
In this paper, we introduced HyperCOD, the first large-scale benchmark for hyperspectral camouflaged object detection, featuring 350 high-resolution hyperspectral images across various camouflage scenarios and environmental conditions. We also proposed HSC-SAM, a spectral-aware adaptation of the Segment Anything Model that integrates spectral-spatial decomposition and saliency-guided token filtering to enhance object-relevant features, achieving state-of-the-art performance. We believe that HyperCOD, along with the proposed framework, will drive significant advancements in hyperspectral camouflaged object detection research.

\bibliography{aaai2026}

\end{document}